%% file: main.tex
\documentclass[10pt,twocolumn,letterpaper]{article}

\usepackage[pagenumbers]{cvpr} 


\definecolor{cvprblue}{rgb}{0.21,0.49,0.74}
\usepackage[pagebackref,breaklinks,colorlinks,allcolors=cvprblue]{hyperref}
\usepackage{multirow}
\usepackage{graphicx}
\usepackage{booktabs}
\usepackage{colortbl}
\usepackage{xcolor}
\usepackage{pifont}
\usepackage{bbding}
\usepackage{float}
\usepackage{colortbl}
\usepackage{xcolor}
\newcommand{\best}[1]{\cellcolor{red!40}{#1}}
\newcommand{\second}[1]{\cellcolor{orange!30}{#1}}
\newcommand{\third}[1]{\cellcolor{yellow!30}{#1}}

\def\paperID{ } 
\def\confName{CVPR}
\def\confYear{2025}

\title{Car-GS: Addressing Reflective and Transparent Surface Challenges in 3D Car Reconstruction}

\author{
Congcong Li, 
Jin Wang, 
Xiaomeng Wang, 
Xingchen Zhou, 
Wei Wu, \\
Yuzhi Zhang, 
and Tongyi Cao\textsuperscript{\Envelope} \\
DeepRoute.AI \\
{\tt\small \{congcongli, jinwang03, xiaomengwang, xingchenzhou, } \\
{\tt\small weiwu, yuzhizhang, tongyicao\}@deeproute.ai}
}

\begin{document}
\maketitle
\input{sec/abstract}
\input{sec/intro}
\input{sec/related_work}
\input{sec/method}

\input{sec/experiment}

\input{sec/conclusion}

\small \bibliographystyle{ieeenat_fullname} \bibliography{main}

\end{document}

%% file: sec/abstract.tex
\begin{abstract}

3D car modeling is crucial for applications in autonomous driving systems, virtual and augmented reality, and gaming. However, due to the distinctive properties of cars, such as highly reflective and transparent surface materials, existing methods often struggle to achieve accurate 3D car reconstruction.
To address these limitations, we propose Car-GS, a novel approach designed to mitigate the effects of specular highlights and the coupling of RGB and geometry in 3D geometric and shading reconstruction (3DGS). Our method incorporates three key innovations: First, we introduce view-dependent Gaussian primitives to effectively model surface reflections. Second, we identify the limitations of using a shared opacity parameter for both image rendering and geometric attributes when modeling transparent objects. To overcome this, we assign a learnable geometry-specific opacity to each 2D Gaussian primitive, dedicated solely to rendering depth and normals. Third, we observe that reconstruction errors are most prominent when the camera view is nearly orthogonal to glass surfaces. To address this issue, we develop a quality-aware supervision module that adaptively leverages normal priors from a pre-trained large-scale normal model. 
Experimental results demonstrate that Car-GS achieves precise reconstruction of car surfaces and significantly outperforms prior methods.
The project page is available at \href{https://lcc815.github.io/Car-GS}{Car-GS}.

\end{abstract}

%% file: sec/intro.tex
\section{Introduction}

\begin{figure}[t]  
  \centering
   \includegraphics[width=1.0\linewidth]{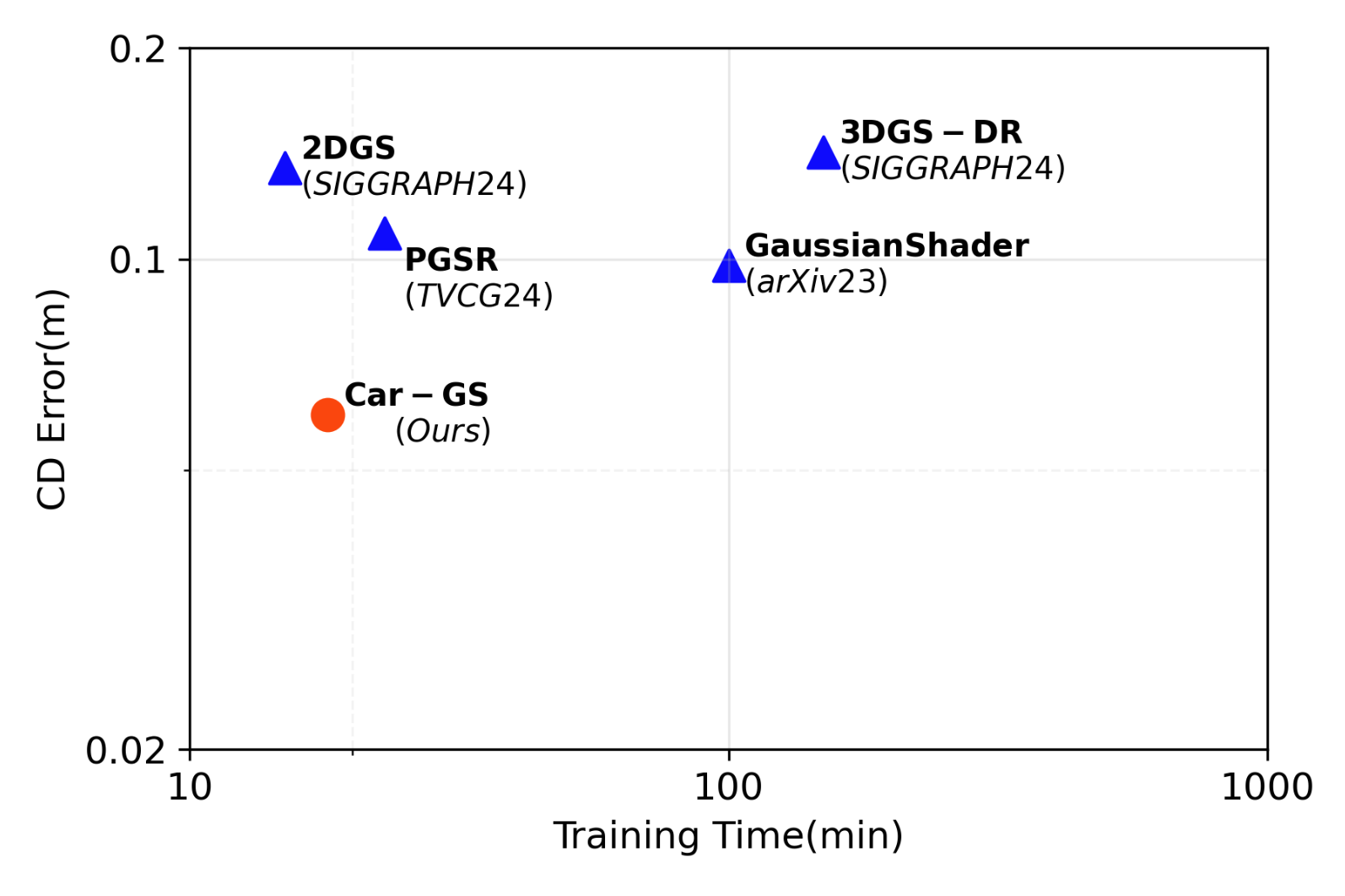}
   \caption{Comparison of various methods based on training time and chamfer distance(CD) error. Our approach not only achieves the highest accuracy but also demonstrates a relatively short training time, highlighting its balance between superior performance and efficiency. This makes it well-suited for real-time applications.}
   \label{fig:teaser_axis}
\end{figure}

\begin{figure*}[t!]  
  \centering
   \includegraphics[width=1.0\linewidth]{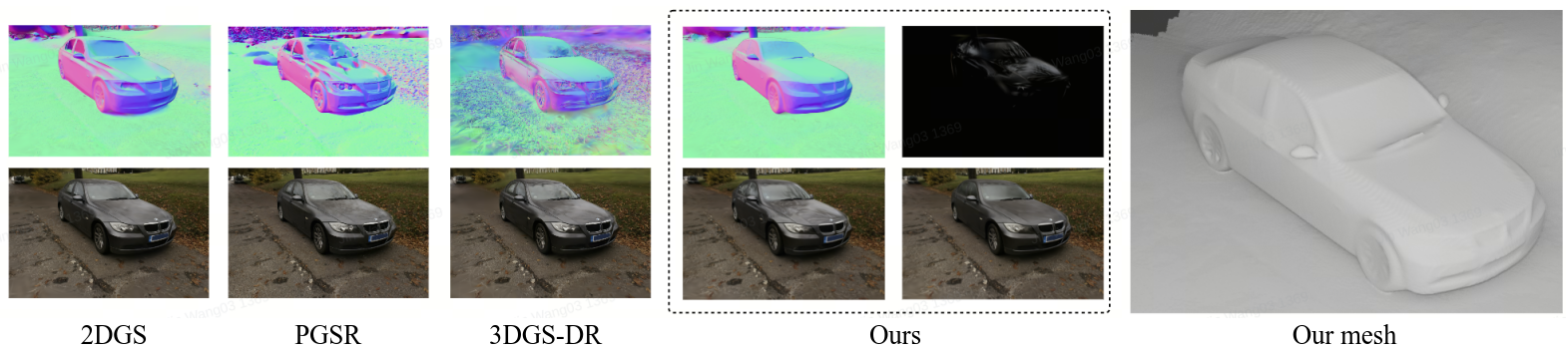}
   \caption{Our Car-GS method accurately separates reflections and recovers surface normals, achieves photo-realistic rendering, and demonstrates superior reconstruction quality of the car geometry, including detailed handling of reflective and transparent regions such as the car's body and windshield.}
   \label{fig:teaser}
\end{figure*}

Accurate 3D reconstruction is a critical task in various domains, including autonomous driving simulation and virtual/augmented reality (VR/AR) applications. However, reconstructing 3D models of cars presents significant challenges due to the complex surface properties of vehicles. Car surfaces are often highly reflective, such as glossy paints, and frequently transparent, such as windows and windshields. These properties pose considerable difficulties in accurately recovering both geometry and texture during 3D reconstruction.

Existing methods, such as NeRF~\cite{mildenhall2021nerf} and its extensions, including Mip-NeRF~\cite{barron2021mip}, Mip-NeRF 360~\cite{barron2022mip}, and NeuS~\cite{wang2021neus}, represent 3D scenes as sets of emission radiance points and compute view-dependent colors based on the viewing direction. However, these approaches do not account for the full interaction of light rays as they travel from the source to the camera, neglecting phenomena such as light scattering and reflection. To address these limitations, Ref-NeRF~\cite{verbin2022ref} incorporates surface light field rendering~\cite{chen2018deep, wood2023surface} and replaces NeRF’s directional parameterization with an integrated reflection encoding. This modification significantly enhances the realism and accuracy of specular reflections.

Recent advancements, such as feature grid-based encoding for the directional domain~\cite{wu2024neural}, have significantly enhanced the efficiency of directional representations. In the context of 3D Gaussian Splatting (3DGS)~\cite{kerbl20233d}, directly applying view direction reflections as view-dependent color queries within efficient Gaussian splatting frameworks introduces several challenges. Specifically, each primitive in the model independently inherits both orientation and spherical harmonics (SH) color, which can result in misalignment when the viewing direction changes during parameter updates.
To address this issue, Spec-Gaussian~\cite{yang2024spec} incorporates smooth regularization and higher-order view-dependent color modeling into the rendering process, thereby achieving notable improvements in rendering reflective surfaces. Nevertheless, despite these advancements in rendering quality, accurately modeling geometry remains a significant challenge.

In this paper, we introduce Car-GS, a novel approach that enhances the reconstruction of vehicles by mitigating the impact of reflective and transparent surfaces. Specifically, we propose using view-dependent Gaussians to model surface reflections. Furthermore, we identify that a crucial limitation of existing methods for modeling transparent surfaces is the shared opacity parameter used for rendering both image and geometric attributes. To overcome this limitation, we assign a learnable, geometry-specific opacity parameter to each 2D Gaussian primitive, which is used exclusively for rendering depth and normals, thereby ensuring a more accurate representation of transparent surfaces. In addition, we incorporate a pre-trained normal model~\cite{ye2024stablenormal} to provide additional geometry supervision during the training process. However, predictions of this model may not be perfect, and we observe that errors are most prominent when the view ray is nearly orthogonal to glass surfaces. Therefore, we develop a quality-aware supervision module that can adaptively refine the rendered normals, leading to improved reconstruction accuracy.

In summary, we make the following contributions: 
\begin{itemize} 
\item We propose the use of view-dependent Gaussians to model surface reflections, thereby enhancing the accuracy of 3D car reconstruction by effectively addressing the challenges posed by reflective surfaces. 
\item We introduce a learnable, geometry-specific opacity parameter for each 2D Gaussian primitive, dedicated solely to rendering depth and normals. This approach overcomes the limitations of using a shared opacity parameter for both image and geometric attributes in existing methods, ensuring a more precise representation of transparent surfaces. 
\item We develop a quality-aware supervision module that leverages normal priors from a pre-trained, large-scale normal model. This module adaptively refines the rendered normals, thereby improving reconstruction accuracy by mitigating errors in normal predictions. 
\item Our approach achieves state-of-the-art performance in car surface reconstruction, outperforming existing methods in terms of accuracy and robustness. 
\end{itemize}

%% file: sec/related_work.tex
\section{Related Work}
\label{sec:related_work}

\begin{figure*}[t!]
  \centering
   \includegraphics[width=1.0\linewidth]{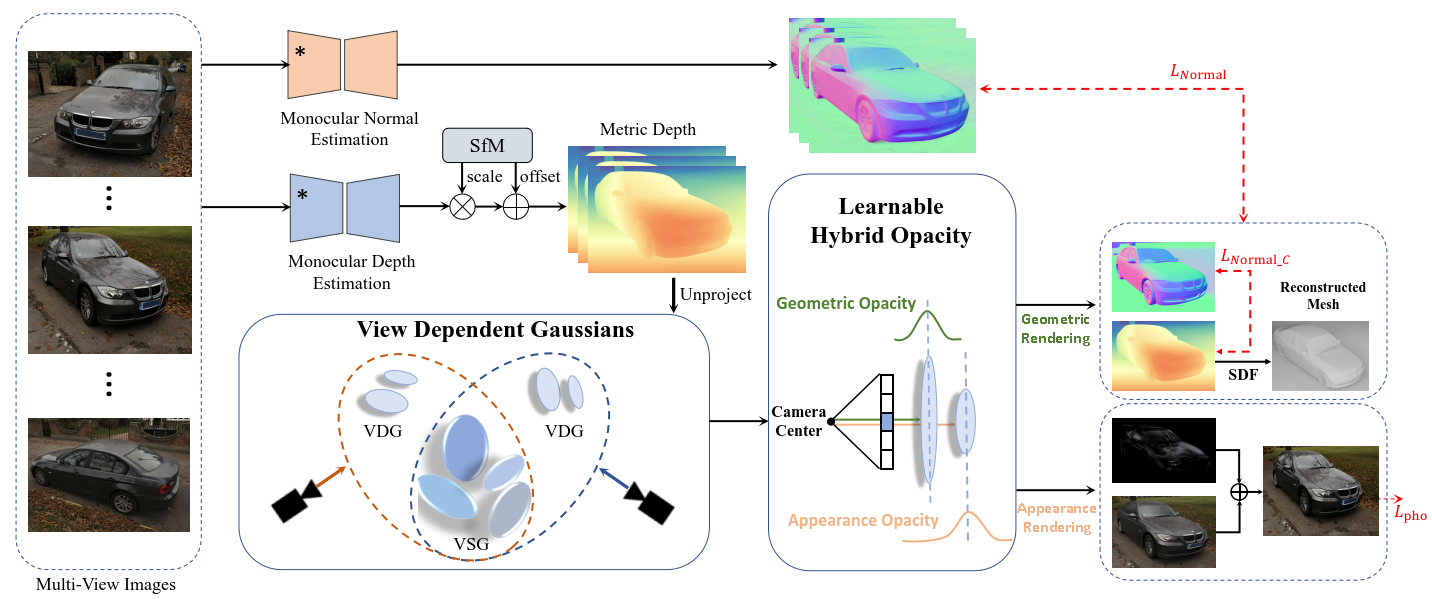}
   \caption{\textbf{Overview of Car-GS.} We initialize View-Dependent Gaussians (VDG) and View-Shared Gaussians (VSG) using monocular depth estimates aligned with structure-from-motion (SfM). VDG models view-specific attributes, while VSG captures shared information. During rendering, a learnable hybrid opacity is applied to RGB images and depth/normal maps. Additionally, a quality-aware supervision module leverages pretrained normal priors to regulate rendered normals, especially in reflective regions.}
   \label{fig:method}
\end{figure*}

\subsection{Novel View Synthesis}
Recent advances in neural implicit representations, particularly Neural Radiance Fields (NeRF)~\cite{mildenhall2021nerf, cheng2023uc, chang2023depth, hu2023tri, liu2024rip}, have revolutionized the field of novel view synthesis. The emergence of NeRF has catalyzed numerous architectural innovations and optimization techniques in this domain. 
Instant-NGP~\cite{muller2022instant} significantly enhanced computational efficiency through the integration of hash-based encoding and adaptive sampling strategies, demonstrating orders of magnitude improvement in both training and inference latency.
TensorIR~\cite{feng2023tensorir} extended the classical NeRF framework by incorporating tensor decomposition techniques for inverse rendering, while MegaNeRF~\cite{turki2022mega} introduced a scalable block-based architecture capable of handling large-scale scene reconstruction.

The introduction of 3D Gaussian Splatting (3DGS) and its variants~\cite{kerbl20233d, lu2024scaffold, ren2024octree, yu2024gaussian, cheng2024gaussianpro} marked a paradigm shift from implicit to explicit scene representations. 
This approach models scenes using a set of 3D Gaussian primitives with associated spherical harmonic coefficients, enabling efficient rasterization-based rendering without the computational overhead of volumetric sampling. 
Although 3DGS achieves superior visual fidelity and temporal coherence compared to NeRF-based methods, it encounters challenges in maintaining geometric consistency across multiple views. 
The subsequent development of 2D Gaussian Splatting (2DGS)~\cite{huang20242d} addressed these limitations by constraining the Gaussian primitives to disk-like structures, effectively reducing the degrees of freedom in the geometric representation. However, 2DGS exhibits notable limitations in accurately reconstructing scenes with specular surfaces and complex reflective properties.

\subsection{Surface Reconstruction}
\textbf{Surface Reconstruction of General Object.} Surface reconstruction of general object has been a long-standing
goal in computer vision. Several NeRF-based reconstruct object surface by incorporating intermediate representations such as
occupancy~\cite{niemeyer2020differentiablevolumetricrenderinglearning} or signed distance fields ~\cite{wang2021neus, yariv2021volumerenderingneuralimplicit}. Although NeRF-based frameworks demonstrate powerful surface reconstruction capabilities, the use of stacked multi-layer perceptron (MLP) layers limits both inference speed and representational capacity. To overcome these limitations, subsequent research has focused on reducing reliance on MLP layers by decomposing scene information into more separable structures, such as points~\cite{xu2022point} and voxels~\cite{li2022vox, li2023neuralangelo, liu2020neural}.
Recently, 3DGS-based surface reconstruction methods ~\cite{Huang2DGS2024, PGSR, zhang2024gspull} have seen rapid development. To solve the problem of difficulty in accurately reconstructing surfaces, some approaches focus on reducing 3D Gaussians to 2D Gaussians by applying a set of regularization terms, ensuring that the Gaussian primitives align with the object surfaces~\cite{Huang2DGS2024, guedon2023sugar,Dai2024GaussianSurfels}. Additionally, some methods incorporate priors from large-scale datasets~\cite{turkulainen2024dnsplatter} or multi-view stereo techniques~\cite{wolf2024gsmesh}, or employ specialized surface extraction algorithms~\cite{Yu2024GOF, ye2024gaustudio} to recover 3D geometry from 3D Gaussians. However, these methods are difficult to reconstruct reflective and transparent object surfaces.

\textbf{Surface Reconstruction of Reflective and Transparent Object.} Objects with strong reflective properties present a significant challenge for surface reconstruction using methods such as vanilla NeRF~\cite{mildenhall2021nerf} and NeuS~\cite{wang2021neus}. This challenge arises from the high-frequency variations in reflections that occur across different viewing angles, resulting in view-dependent color inconsistencies at the same surface point, a phenomenon known as anisotropy. To overcome these limitations, various approaches have been proposed to enhance reflective scene reconstruction within the NeRF framework. Notable examples include methods like Ref-NeRF~\cite{verbin2022ref}, NeRF-Casting~\cite{verbin2024nerf}, and inverse rendering techniques such as NeRO~\cite{liu2023nero}, which specifically address the complexities of anisotropy to improve reconstruction accuracy in reflective environments.

Ref-NeRF~\cite{verbin2022ref} modified NeRF's original formulation by replacing outgoing radiance with reflective radiance, introducing view-dependent scene attributes to decouple specular components. NeRF-Casting~\cite{verbin2024nerf} employed fundamental ray-tracing principles, sampling along reflection rays to achieve high-quality reconstruction of scenes containing reflective objects. While NeuS-based methods are generally better suited for surface representation, their inherent isotropic mesh representation poses significant challenges when dealing with complex reflective scenes, resulting in notable artifacts.
UniSDF~\cite{wang2023unisdf} introduced a learnable weighting network to combine camera view radiance fields with reflected view radiance fields. Their approach, rather than directly modeling reflective and non-reflective regions separately, demonstrated superior results in view-dependent mesh reconstruction. Ref-NeuS~\cite{ge2023ref} achieved high-quality reflective surface reconstruction by incorporating reflection-aware photometric loss and reflection direction-dependent radiance to reduce ambiguities caused by reflective surfaces.
Gaussian Shader~\cite{Gaussianshader}, built upon 3D Gaussian Splatting, presents a novel approach to modeling both reflective and non-reflective regions. It employs simplified shading functions to decouple the modeling of reflective scenes, offering an effective solution for handling complex reflective environments.

%% file: sec/method.tex
\section{Method}

\begin{table*}[t!]
\centering
\renewcommand{\arraystretch}{1.2}
\begin{tabular}{@{}lccccccccc@{}}
\toprule
\multirow{2}{*}{\textbf{Method}} & \multicolumn{5}{c}{\textbf{3DRealCar}}                  & \multirow{2}{*}{\textbf{Mean}} & \multirow{2}{*}{\textbf{Training Time}} \\ \cmidrule(lr){2-6}
                                 & Scene 1   & Scene 2   & Scene 3   & Scene 4   & Scene 5   &                              &                                         \\ \midrule
\multicolumn{8}{c}{\textbf{Chamfer Distance $\downarrow$}} \\ \midrule
3DGS-DR                          & 0.156     & 0.120     & 0.147     & 0.169     & 0.120     & 0.142                        & 2h30min                                 \\
Gaussian Shader                  & \second{0.073} & \third{0.077} & \second{0.105}     & \second{0.136}     & \third{0.099} & \second{0.098}              & ~1h40min                                \\
PGSR                             & \third{0.115}  & \second{0.064} & \third{0.127}     & \third{0.142} & \second{0.095} & \third{0.109}               & \third{~23min}                          \\
2DGS                             & 0.134     & 0.104     & 0.133     & 0.190     & 0.115     & 0.135                        & \best{~15min}                           \\
\midrule
Ours                             & \best{0.067} & \best{0.052} & \best{0.059} & \best{0.060} & \best{0.062} & \best{0.060}               & \second{~20min}                         \\ \midrule \midrule
\multicolumn{8}{c}{\textbf{Accuracy $\uparrow$}} \\ \midrule
3DGS-DR                          & 0.314      & 0.460       & 0.348      & 0.343      & 0.457      & 0.384                         & 2h30min                                 \\
Gaussian Shader                  & \second{0.723} & \third{0.707} & 0.483     & \third{0.468} & \third{0.566} & \third{0.589}               & ~1h40min                                \\
PGSR                             & 0.473      & \second{0.794} & \second{0.550} & \second{0.550} & \second{0.635} & \second{0.600}                 & \third{~23min}                          \\
2DGS                             & \third{0.576} & 0.675     & \third{0.541} & 0.314      & 0.537      & 0.529                         & \best{~15min}                           \\
\midrule
Ours                             & \best{0.745} & \best{0.825} & \best{0.740}  & \best{0.636} & \best{0.808} & \best{0.751}               & \second{~20min}                         \\ \midrule \midrule
\multicolumn{8}{c}{\textbf{F1 Score $\uparrow$}} \\ \midrule
3DGS-DR                          & 0.443      & 0.596      & 0.464      & 0.464      & 0.571      & 0.508                         & 2h30min                                 \\
Gaussian Shader                  & \best{0.817} & \third{0.783} & 0.581     & \third{0.560}       & \third{0.661}      & \second{0.680}                & ~1h40min                                \\
PGSR                             & 0.567      & \second{0.830}  & \third{0.609}     & \second{0.644}      & \second{0.713}      & \third{0.673}                & \third{~23min}                          \\
2DGS                             & \third{0.687}  & \third{0.768}  & \second{0.625}  & 0.401  & 0.629  & 0.622                         & \best{~15min}                           \\
\midrule
Ours                             & \second{0.776} & \best{0.833} & \best{0.745}  & \best{0.664}  & \best{0.800}    & \best{0.764}               & \second{~20min}                         \\ \bottomrule \bottomrule
\end{tabular}
\caption{Quantitative results of reconstruction quality on the 3DRealCar dataset. 'Red', 'Orange', and 'Yellow' denote the best, second-best, and third-best results, respectively. Our proposed method outperforms existing mesh reconstruction techniques in terms of reconstruction quality, while also achieving relatively shorter training times.}
\label{tab:3drealcar_metrics}
\end{table*}

\begin{figure}[t!]
  \centering
   \includegraphics[width=1.0\linewidth]{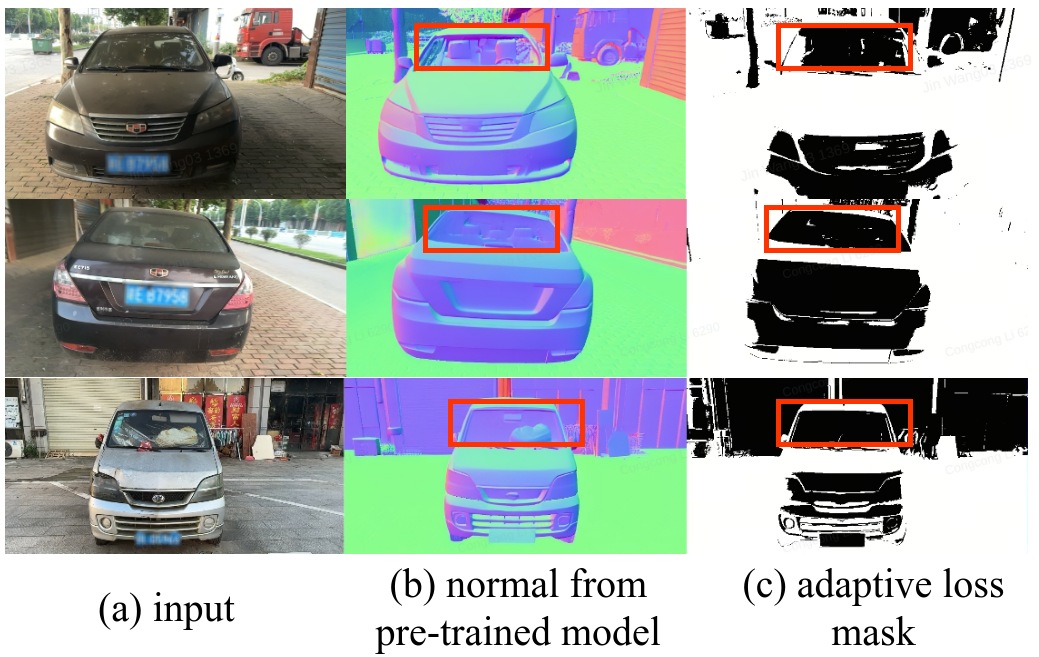}
   \caption{(a), (b), and (c) represent the input image, the normal map from the pretrained model~\cite{ye2024stablenormal}, and the loss mask computed by~\cref{eq:qam}, respectively. In the mask, black indicates a value of 0, while white indicates a value of 1. We observe that errors in the pretrained model's predictions are most prominent when the camera view is nearly orthogonal to the glass surface. Therefore, we adaptively assign a value of 0 to these orthogonal regions.} 
\label{fig:normal_error}
\end{figure}

Our approach is based on 2DGS~\cite{Huang2DGS2024} and we aim to accurately reconstruct the surfaces of cars. An overview of our method is shown in~\cref{fig:method}. Specifically, we first present view-dependent Gaussians to model reflections from each view. Next, we introduce a novel learnable hybrid opacity to reduce ambiguity during the rendering of both appearance and geometry maps. Finally, we introduce a quality-aware supervision module to adaptively refine rendered normals.

\subsection{View-Dependent Gaussians}
\label{sec:View-Dependent Gaussians}

In computer graphics, the color of an object's surface, $I_{\text{obj}}$, is equal to the sum of ambient light $I_{\text{ambient}}$, diffuse reflection $I_{\text{diffuse}}$, and specular reflection $I_{\text{specular}}$:
\begin{equation}
I_{\text{obj}} = I_{\text{ambient}} + I_{\text{diffuse}} + I_{\text{specular}}.
\end{equation}
Among these components, specular reflection is view-dependent, which means that the presence of specular reflection may lead to different observations of the same point in 3D space from different views and cause ambiguities and ultimately affect the quality of geometric surface reconstruction. Therefore, we first aim to remove the specular reflection from each viewpoint, leaving only the view-independent components: ambient light and diffuse reflection. For each view, we define a set of Gaussians, termed view-dependent Gaussians (VDG), and train them to fit the specular highlights for the current viewpoint. During the evaluation and mesh extraction process, these VDG are discarded, and only the view-shared Gaussians (VSG) are involved.

\textbf{Initialization of VDG}.
Unlike the original 3DGS, VDG are designed to represent reflections. These reflections often occur in texture-less regions, which are typically absent from the initial point cloud generated by COLMAP. To address this limitation, we use a point cloud converted from an estimated monocular depth map as our initial VDG. Specifically, we first utilize an open source instance segmentation model SAM2~\cite{ravi2024sam2} to obtain the mask of the car. Within the mask, a set of pixels is randomly sampled. Based on the camera's intrinsic matrix $\mathbf{K}$, each sampled pixel $[x_i, y_i]$ (the pixel coordinates in the image plane) is mapped to a corresponding point $\mathbf{p}_{\text{camera}, i}$ in the camera coordinate system:
\begin{equation}
\mathbf{p}_{\text{camera}, i} = \mathbf{K}^{-1} \begin{bmatrix}
 x_i \cdot z_i \\
 y_i \cdot z_i \\
 z_i
\end{bmatrix}.
\end{equation}
During this mapping process, the depth value $z_i$ for each sampled pixel is obtained from monocular depth estimation~\cite{depth_anything_v2} result $z_{rel,i}$, which represents relative depth. This depth is then aligned with COLMAP's scale:
\begin{equation}
z_i = z_{rel,i} \cdot s + o,
\end{equation}
where $s$ and $o$ represent the scale and offset, respectively, obtained following the method in 3DGS~\cite{kerbl20233d}. The mapped points are subsequently transformed into the world coordinate system, serving as the initialization points for the VDG:
\begin{equation}
\mathbf{p}_{\text{world}, i} = \mathbf{R} \cdot \mathbf{p}_{\text{camera}, i} + \mathbf{T}.
\end{equation}
where 
$R$ and $T$ represent the rotation matrix and translation vector, respectively, for transforming coordinates from the camera frame to the world frame.

\textbf{Regularization of VDG}.
Training these VDG in the same manner as VSG does not yield the desired results, as these VDG may overfit the current view. To mitigate this, we incorporate opacity regularization into the loss function, ensuring that the VDG contribute to the rendered image only when necessary. Specifically, the loss is defined as:
\begin{equation}
\mathcal{L}_{{vdg}} = w_{{vdg}} \frac{\sum_{i=1}^{N_{vdg}}  o_{{vdg, i}}}{N_{vdg}},
\end{equation}
where $N_{vdg}$ is the total number of VDG, $o_{{vdg,i}}$ is the opacity of the $i$-th VDG and $w_{vdg}$ is the loss weight.

\begin{figure*}[t!]
  \centering
   \includegraphics[width=1.0\linewidth]{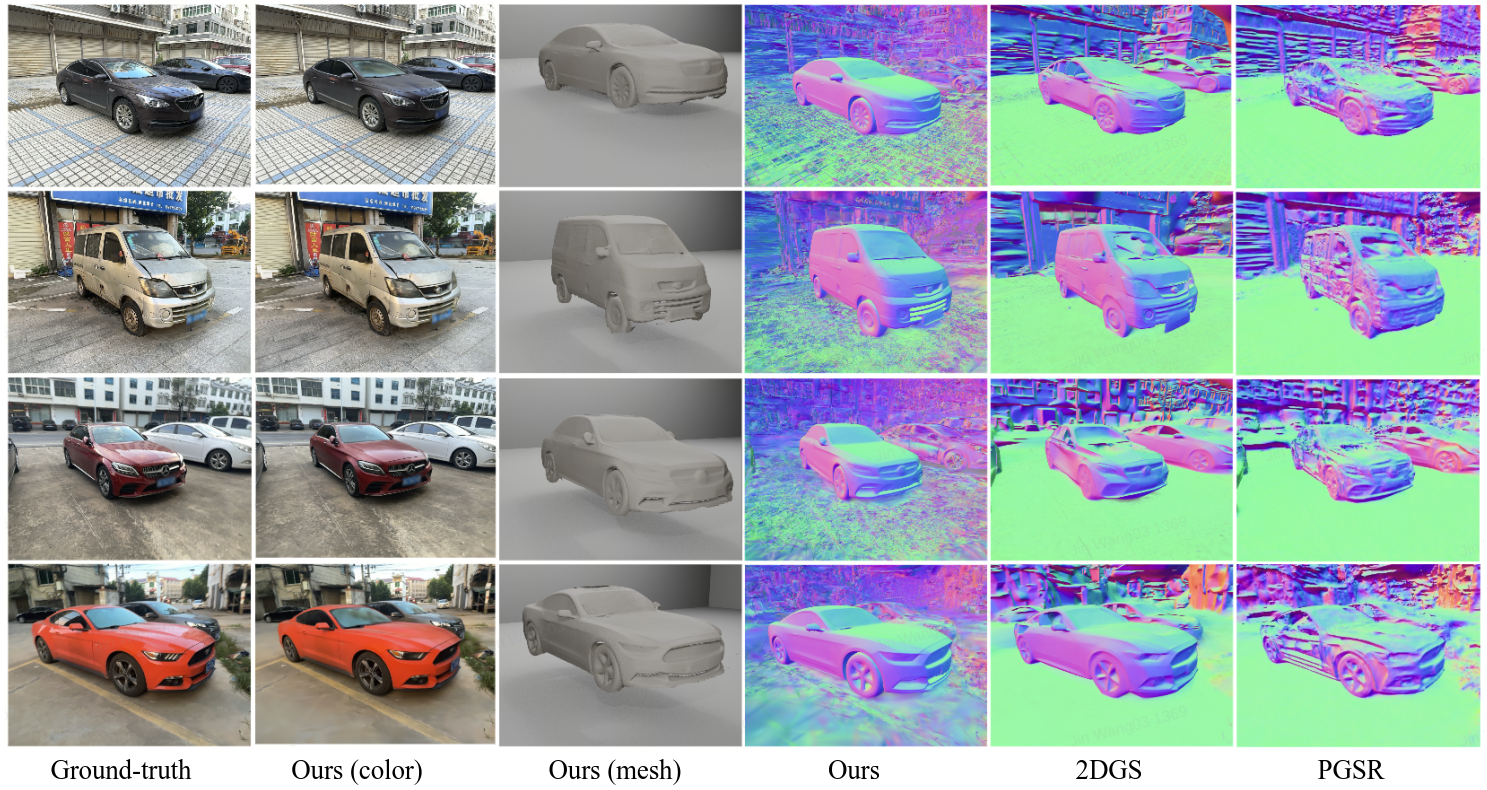}
   \caption{Visual comparisons on test-set views from the 3DRealCar dataset. Note that we focus on the reconstruction of the vehicle body rather than the entire scene. Our method excels at synthesizing geometrically accurate radiance fields and surface reconstructions, outperforming other baseline approaches in capturing sharp edges and intricate details. In contrast, baseline models often fail in areas with transparent glass and reflective surfaces on the vehicle body.}
   \label{fig:3drealcar_comparison}
\end{figure*}

\subsection{Learnable Hybrid Opacity}
\label{sec:Learnable Hybrid Opacity}
Current GS-based methods use the same opacity parameter for rendering both appearance and geometry maps. However, for transparent objects such as glass, depth rays and appearance rays should behave differently. The depth ray should terminate at the object's surface, whereas the appearance ray should continue to propagate forward.

To address this issue, we propose a learnable hybrid opacity (LHO), which includes two parameters: appearance opacity (sometimes simply referred to as opacity in this paper) and geometric-specific opacity. These parameters are used to separately render RGB images and geometry maps, e.g., normal and depth maps. The rendered map $M$ is computed as follows:
\begin{equation}
M = \sum_{i \in N} attr_i \alpha_i \prod_{j=1}^{i-1} (1 - \alpha_j),
\end{equation}
where $\text{attr}_i$ represents the color of the $i$-th Gaussian when $M$ is an RGB map, and $\alpha_i$ is derived by evaluating the 2D covariance matrix $\Sigma$~\cite{Yifan:DSS:2019}, scaled by the appearance opacity. Similarly, $\text{attr}_i$ represents the geometric attribute of the $i$-th Gaussian when $M$ is a geometry map, and $\alpha_i$ is computed by evaluating $\Sigma$, scaled by the geometry opacity.

Based on our LHO, transparent surfaces can be modeled by learning the geometry opacity as $1$ and the RGB opacity as $0$, for example. However, such a design provides the model with greater flexibility, which could lead to overfitting. Specifically, We use a loss item similar to 2DGS to align the splats’ normal $N_s$ with the gradients of the depth maps $N_d$ as follows:
\begin{equation}
\mathcal{L}_{ds} = w_{ds}|N_d - N_s|,
\end{equation}
where $w_{ds}$ is the corresponding loss weight. However, when there is a discrepancy between the geometry opacity and appearance opacity, the model may learn completely transparent points at arbitrary locations to reduce the aforementioned loss terms without affecting other loss components. This would ultimately result in unpredictable reconstruction outcomes. Therefore, we constrain the geometry opacity and appearance opacity to be as consistent as possible. Specifically, we introduce another loss item:
\begin{equation}
\mathcal{L}_{lho} = w_{lho} \frac{\sum_{k=1}^{N} |o_k - geo\_o_k|}{N},
\end{equation}
where $w_{lho}$ is the loss weight, $N$ is the total number of view-shared Gaussians, $o_k$ and $geo\_o_k$ are the appearance opacity and geometry opacity respectively of the $k$-th Gaussian.

Note that, since our VDG are used for modeling reflections, we do not assign LHO to these Gaussians. In addition, in the original 2DGS, the appearance opacity of each 2D primitive is reset every few iterations and may be pruned due to its small opacity. In this work, we simultaneously reset the hybrid opacity, and prune the 2D Gaussian only when both the appearance opacity and the geometry opacity are small.

\subsection{Quality-aware Supervised Module}
\label{sec:Pixel-wise Ground Truth Quality-aware Module}

In texture-less regions, adding more views does not provide supplementary geometric information. We argue that reconstruction of these surfaces using only RGB information is an ill-posed problem. To address this, we introduce additional supervision based on surface normals. 

Specifically, we employ an open-source, pre-trained normal estimation model~\cite{ye2024stablenormal} to serve as pseudo labels $\hat{n}$, which are used to constrain the normals of VSG. Therefore, our final normal loss function is defined as follows:
\begin{equation}
    \mathcal{L}_{\text{normal}} = w_n|N_d - {\hat{n}}| + w_n|{N_s} - {\hat{n}}| + \mathcal{L}_{ds},
\end{equation}
where $w_{n}$ is a weighting factor, which is determined dynamically by our pixel-wise quality-aware supervision module(QSM).

This quality-aware supervision module is designed to address the imperfections inherent in predictions made by pre-trained models. We observe that such errors predominantly occur when the view ray is nearly orthogonal to the 2D Gaussian, as shown in \cref{fig:normal_error} (b).
Based on this observation, we calculate the angle $\theta$ between the normal vector of each 2D Gaussian and the pixel ray. If this angle value falls below a certain threshold $\tau$, we disregard the normal supervision for that pixel, as it likely corresponds to an incorrect pseudo label. Formally, $w_n$ is defined as:
\begin{equation}
    w_n = \begin{cases}
        1, & \text{if } \theta > \tau, \\
        0, & \text{if } \theta \leq \tau.
    \end{cases}
\label{eq:qam}
\end{equation}

This approach ensures that the supervision is applied adaptively, enhancing the robustness of the reconstruction process in challenging regions with incorrect normal predictions, as shown in \cref{fig:normal_error} (c), where it can be observed that after incorporating our proposed quality-aware supervision module, we adaptively set the constraint weight to zero for these incorrect regions, effectively mitigating the impact of erroneous priors.

\begin{figure*}[t!]
  \centering
   \includegraphics[width=1.0\linewidth]{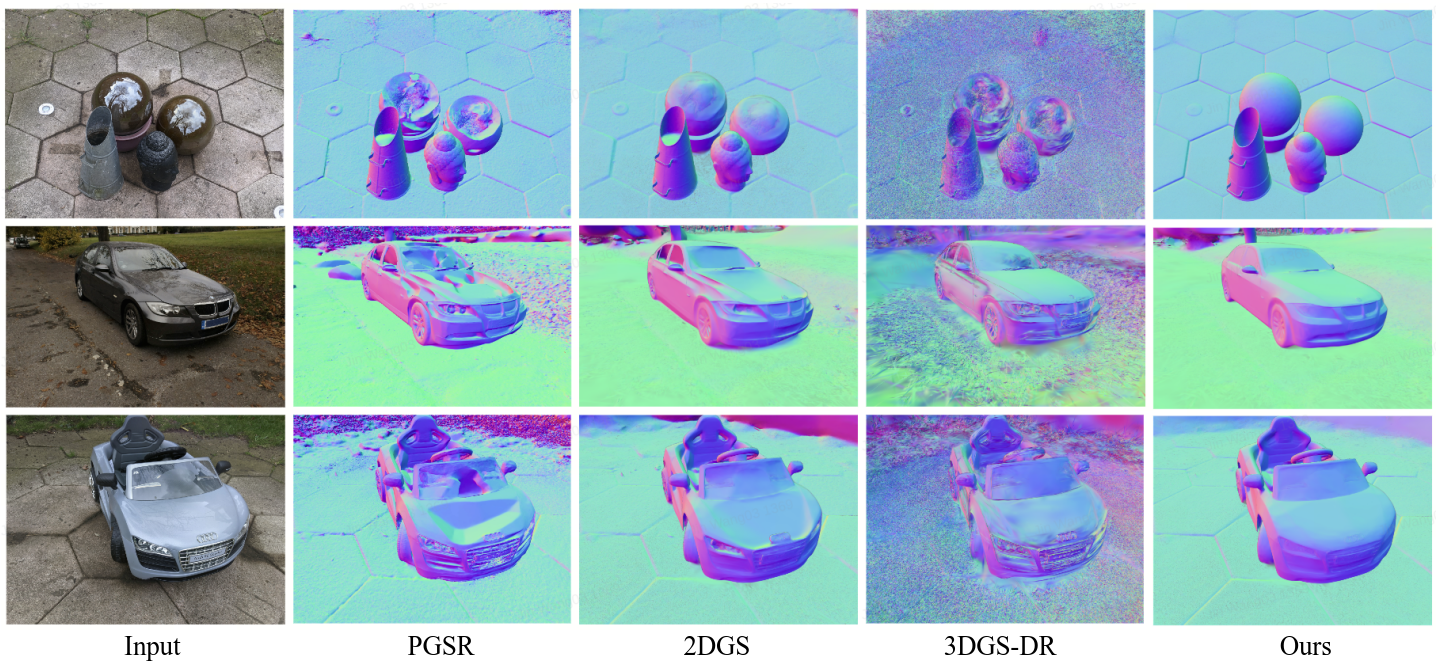}
   \caption{Visual comparisons on test-set views from the Ref-Nerf real dataset. Our method achieves superior normal estimations, particularly in regions with reflective surfaces and glass, where other approaches fail and exhibit missing or noisy surfaces.}
   \label{fig:refnerf_comparison}
\end{figure*}

\subsection{Loss Functions}
\label{sec:LossFunctions}

Our total loss function, which integrates multiple components to optimize the reconstruction process, is defined as follows:
\begin{equation}
    \mathcal{L}_{{total}} = \mathcal{L}_{c} + \mathcal{L}_{{normal}} + \mathcal{L}_{{vdg}} + \mathcal{L}_{lho},
\end{equation}
where $\mathcal{L}_{c}$ represents the RGB reconstruction loss. This loss combines the $\ell_1$ loss and the D-SSIM term, as proposed in 3DGS~\cite{kerbl20233d}, ensuring high-fidelity color and structural consistency in the reconstructed images.

The weighting factors for the individual loss components are set as $w_{ds} = w_{n} = 0.1$ in $\mathcal{L}_{{normal}}$, $w_{{vdg}} = 0.2$ for the VDG-related term $\mathcal{L}_{{vdg}}$, and $w_{{lho}} = 3.0$ for the term of LHO optimization $\mathcal{L}_{{lho}}$.

%% file: sec/experiment.tex
\section{Experiments}

\begin{table*}[ht]
    \centering
    \begin{tabular}{l|ccc|ccc|ccc}
        \toprule
        \multirow{2}{*}{Method} & \multicolumn{3}{c|}{3DRealCar} & \multicolumn{3}{c|}{Ref-NeRF} & \multicolumn{3}{c}{Average} \\
        \cmidrule{2-10}
         & PSNR $\uparrow$ & SSIM $\uparrow$ & LPIPS $\downarrow$ & PSNR $\uparrow$ & SSIM $\uparrow$ & LPIPS $\downarrow$ & PSNR $\uparrow$ & SSIM $\uparrow$ & LPIPS $\downarrow$ \\
        \midrule
        3DGS-DR 
            & 23.954 
            & 0.767 
            & 0.343 
            & 22.633 
            & 0.585 
            & 0.382 
            & 23.294 
            & 0.676 
            & 0.363 \\
        
        Gaussian Shader 
            & 23.246 
            & 0.767 
            & {0.326} 
            & 23.463 
            & 0.647 
            & 0.257 
            & 23.355 
            & 0.707 
            & 0.292 \\
        
        PGSR 
            & \best{25.818} 
            & \best{0.876}
            & \best{0.122} 
            & 23.180 
            & 0.615 
            & 0.319 
            & \best{24.499} 
            & 0.746 
            & \best{0.220} \\
        
        2DGS 
            & \second{24.608} 
            & \second{0.818} 
            & \second{0.242} 
            & \best{24.307} 
            & \best{0.679} 
            & \second{0.279} 
            & \second{24.458} 
            & \best{0.749} 
            & \second{0.261} \\
        
        Ours 
            & \third{23.982} 
            & \third{0.803} 
            & \third{0.258} 
            & \second{23.843} 
            & \second{0.664} 
            & \third{0.287} 
            & \third{23.913} 
            & \second{0.734} 
            & \third{0.273} \\
        \bottomrule
    \end{tabular}
    \caption{Quantitative comparisons of rendering quality, evaluated using PSNR, SSIM, and LPIPS metrics, on the 3DRealCar and Ref-Nerf real datasets. The results are highlighted with 'Red' indicating the best performance, 'Orange' the second-best, and 'Yellow' the third-best.}
    \label{tab:rendering_comparison}
\end{table*}

\subsection{Datasets and Evaluation Metrics}
\label{sec:DatasetsMetrics}
We conducted experiments on the 3DRealCar~\cite{du20243drealcar} and Ref-NeRF Real~\cite{verbin2022ref} datasets. The 3DRealCar dataset represents the first extensive 3D dataset of real cars, comprising over 2,500 vehicles captured under diverse lighting conditions. Each car is represented by an average of 200 dense, high-resolution 360-degree RGB-D views, facilitating high-fidelity 3D reconstruction. Ground truth data for evaluation were derived from meshes obtained using a high-precision 3D scanner.
The Ref-NeRF Real dataset includes three large-scale scenes characterized by reflective surfaces, namely "sedan," "garden spheres," and "toycar." Due to the inherent challenges in accurately capturing glass surfaces with a 3D scanner, we carefully selected five scenes with reliable ground truth data to ensure robustness and fairness in the evaluation process. These scenes were labeled as 20240710062023, 20240707060631, 20240709190817, 20240710052200, and 20240710051501, and are referred to as Scene 1, Scene 2, Scene 3, Scene 4, and Scene 5, respectively, as shown in~\cref{tab:3drealcar_metrics}. To evaluate surface quality, we utilized the Chamfer Distance metric along with Accuracy and F1-score measurements.

Note that our method primarily focuses on the geometric reconstruction of cars, with particular emphasis on accurately capturing challenging surface regions, such as highly reflective areas (e.g., glossy paint) and frequently transparent components. In addition to geometric reconstruction, our method is also capable of delivering high-quality novel view synthesis.
To evaluate the quality of appearance reconstruction, we conducted experiments on the 3DRealCar and the Ref-NeRF real datasets. For the assessment of novel view synthesis, we selected three widely-used metrics: peak signal-to-noise ratio (PSNR), structural similarity index measure (SSIM), and learned perceptual image patch similarity (LPIPS)~\cite{zhang2018unreasonable}. These metrics were employed to validate the fidelity and perceptual quality of the generated novel views.

\subsection{Implementation Details}
We implement our Car-GS with custom CUDA kernels, building upon the framework of 2DGS. Specifically, We use opacity and geometry opacity to render rgb image and geometry maps(normal/depth map), respectively. We initialize the number of VDGs for each camera perspective to 10,000 and keep it constant during the training process and discard these VDGs during test. We conduct all the experiments on a single L20 GPU.

\textbf{Mesh Extraction}. Following 2DGS, we render depth maps of the training views using the depth value of the splats projected to the pixels and utilize truncated signed distance fusion (TSDF) to fuse the reconstruction depth maps, using Open3D. We set the voxel size to 0.004 and the truncated threshold to 0.02 during TSDF fusion. We also extend the original 3DGS-DR to render depth and employ the same technique for surface reconstruction for a fair comparison.

\subsection{Quantitative and Qualitative Results}

\textbf{Results on the 3DRealCar Dataset.}
Quantitative results on reconstruction quality for the 3DRealCar dataset are presented in \cref{tab:3drealcar_metrics}. Compared to state-of-the-art (SOTA) 3D reconstruction methods, especially those designed to handle reflective surfaces, our proposed method demonstrates significant improvements in reconstruction quality. Furthermore, it achieves shorter training times in comparison to existing mesh reconstruction techniques. In \cref{fig:3drealcar_comparison}, we present visual comparisons of the reconstructed normals and meshes generated by both our method and the competing methods. Notably, our approach delivers more detailed mesh reconstructions and more accurate normal estimations, particularly in challenging regions such as reflective surfaces and transparent materials like glass. In contrast, existing methods often introduce substantial noise in these areas, while our approach preserves finer details.

\textbf{Results on the Ref-NeRF Real Dataset.}
Since ground truth geometry is unavailable for the Ref-NeRF real dataset, we provide visual comparison results in \cref{fig:refnerf_comparison}. As observed, our method achieves robust reconstruction results in reflective and transparent glass regions, where competing methods fail. Remarkably, our approach consistently yields geometrically accurate surface reconstructions. Compared to other methods, it achieves significantly superior results in both visual quality and geometric consistency, as evidenced by higher numerical performance metrics and more realistic visual outputs. These results highlight the advantages of our method in handling challenging real-world scenarios, particularly in datasets where ground truth data may not be accessible.

Although our primary focus is on geometric reconstruction, we also present a comparison of rendering performance. Quantitative results are provided in \cref{tab:rendering_comparison}. As shown, in addition to outperforming competing methods in reconstruction quality, our approach also demonstrates superior rendering performance, further emphasizing the efficacy of our method.

\begin{table}[ht]
    \centering
    \resizebox{0.4\textwidth}{!}{%
    \begin{tabular}{l|c|c|c}
        \hline
        Model Setting & CD$\downarrow$ & Accuracy$\uparrow$ & F1 Score$\uparrow$ \\
        \hline
        w/o VDG & 0.096 & 0.513 & 0.574 \\
        w/o LHO & 0.072 & 0.581 & 0.604 \\
        w/o QSM & 0.079 & 0.594 & 0.619 \\
        \textbf{Full Model} & \textbf{0.060} & \textbf{0.636} & \textbf{0.664} \\
        \hline
    \end{tabular}}
    \caption{Ablation studies on the proposed View-Dependent Gaussians (VDG), Learnable Hybrid Opacity (LHO), and Quality-Aware Supervision (QAS) module on the 3DRealCar dataset.}
    \label{tab:ablation_results}
\end{table}

\subsection{Ablation Study} 
\begin{figure}[htbp]
  \centering
   \includegraphics[width=1.0\linewidth]{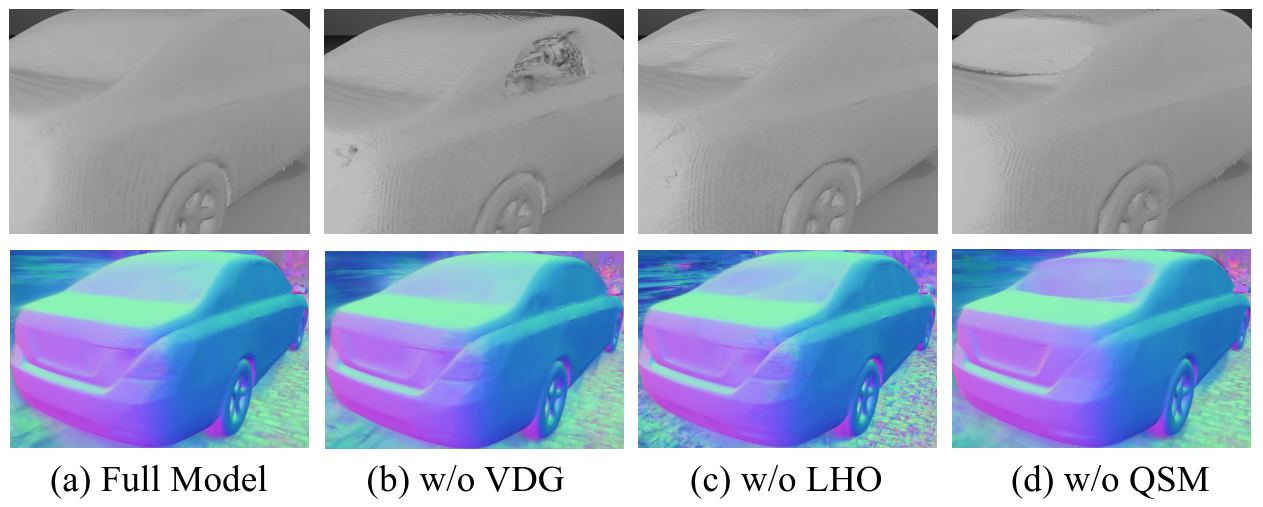}
   \caption{Rendered normal maps and reconstructed meshes on the 3DRealCar dataset, with different components removed. The best performance is achieved when all three components—View-Dependent Gaussians (VDG), Learnable Hybrid Opacity (LHO), and Quality-Aware Supervision (QAS)—are present. Omitting any of these components leads to a progressive degradation in performance.}
   \label{fig:ablation_all}
\end{figure}

In this section, we conduct ablation studies on our model using the 3D real car dataset. Quantitative results are presented in \cref{tab:ablation_results}, and qualitative visualizations are shown in \cref{fig:ablation_all}.

\textbf{Effectiveness of View-Dependent Gaussians.} Specular reflections frequently appear on car surfaces, particularly in metallic and glass regions. These reflections cause the color of the same 3D point to vary across different viewpoints, introducing ambiguity in geometric reconstruction. By incorporating View-Dependent Gaussians (VDGs) to model specular highlights for each viewpoint, our model effectively mitigates the interference of view-dependent color variations on geometry reconstruction. As demonstrated in \cref{tab:ablation_results}, removing VDGs leads to a significant decline in reconstruction accuracy, especially in edge and detail regions. Additionally, the visual results in \cref{fig:ablation_all} illustrate that models without VDGs exhibit noticeable errors and discontinuities in areas with high specular reflections, confirming the crucial role of VDGs in enhancing reconstruction precision.

\textbf{Impact of Learnable Hybrid Opacity.} The introduction of Learnable Hybrid Opacity assigns both geometry opacity and appearance opacity to each Gaussian, enabling more accurate modeling of complex surfaces such as transparent windows. In our ablation experiments, we evaluated the performance of the model without the learnable hybrid opacity mechanism. The results indicate that omitting hybrid opacity leads to inconsistencies between depth and color information in transparent regions, resulting in erroneous surface reconstructions. 
For instance, depth maps may incorrectly penetrate through glass surfaces, or there may be a mismatch between RGB opacity and geometry opacity, causing unnatural visual artifacts. Quantitative metrics in \cref{tab:ablation_results} show that incorporating the learnable hybrid opacity significantly improves the accuracy of depth and normal maps. Moreover, qualitative assessments in \cref{fig:ablation_all} reveal that transparent areas are reconstructed with greater realism and finer details when hybrid opacity is utilized.

\textbf{Effectiveness of Quality-aware Supervised Module.} Reconstructing surfaces with missing textures, such as the body of a car, poses substantial challenges due to the lack of supplementary geometric information from additional views. To address this, we introduced a Quality-aware Supervised Module that adaptively leverages ground truth normals from a pre-trained normal estimation model to provide additional constraints. 
In the ablation study, removing this module resulted in a significant decrease in geometry reconstruction accuracy, leading to loss of geometric details and increased inconsistencies in surface reconstruction. The Quality-aware Supervised Module selectively applies supervision based on the reliability of normal predictions, thereby preventing erroneous normals from degrading reconstruction quality. As shown in \cref{tab:ablation_results}, the inclusion of this module reduces overall normal errors and enriches the geometric details of the reconstructed surfaces. Furthermore, \cref{fig:ablation_all} demonstrates that the module enhances the reconstruction quality in challenging regions with ambiguous or incorrect normal predictions.


%% file: sec/conclusion.tex
\section{Limitation and Conclusion}

In this work, we presented Car-GS, a novel approach designed to tackle the challenges of reconstructing 3D models of vehicles with reflective and transparent surfaces. By introducing view-dependent Gaussians for modeling surface reflections and addressing the key limitation of shared opacity in rendering images and geometric attributes, we significantly improved the accuracy of 3D car surface reconstruction. Our method decouples geometry opacity from RGB opacity, providing a more precise representation of transparent surfaces and mitigating the ambiguity caused by specular highlights. Additionally, we incorporated pretrained large models and proposed a pixel-wise ground truth quality-aware module, which adaptively adjusts loss weights to handle potential inaccuracies in model predictions.
Through extensive experimentation on a recently released real-world vehicle dataset, we demonstrated that Car-GS not only accurately reconstructs car surfaces but also outperforms existing methods. Our results suggest that the proposed approach effectively addresses the unique challenges posed by reflective and transparent materials, offering a promising solution for applications that require high-fidelity 3D reconstructions. Future work can extend this framework to other complex materials and explore further improvements in handling varying levels of transparency and reflectivity, potentially advancing the state-of-the-art in 3D reconstruction for diverse domains.